\documentclass[letterpaper, 10 pt, conference]{ieeeconf}  

\IEEEoverridecommandlockouts                              

\usepackage{amsmath} 
\usepackage{graphicx}
\usepackage{float}
\usepackage{subfigure}
\usepackage{caption}
\usepackage{color}
\usepackage{array}

\usepackage{multirow}
\newcommand{\tabincell}[2]{\begin{tabular}{@{}#1@{}}#2\end{tabular}}

\title{\LARGE \bf
Hierarchical Segment-based Optimization for SLAM
}

\author{Yuxin Tian, Yujie Wang, Ming Ouyang, Xuesong Shi
\thanks{The authors are with Intel Corporation, Beijing 100190, China. Correspondence should be addressed to xuesong.shi@intel.com.}
}

\begin{document}
\maketitle
\thispagestyle{empty}
\pagestyle{empty}
\begin{abstract}


This paper presents a hierarchical segment-based optimization method for Simultaneous Localization and Mapping (SLAM) system. First we propose a reliable trajectory segmentation method that can be used to increase efficiency in the back-end optimization. Then we propose a buffer mechanism for the first time to improve the robustness of the segmentation. During the optimization, we use global information to optimize the frames with large error, and interpolation instead of optimization to update well-estimated frames to hierarchically allocate the amount of computation according to error of each frame. Comparative experiments on the benchmark show that our method greatly improves the efficiency of optimization with almost no drop in accuracy, and outperforms existing high-efficiency optimization method by a large margin.

\end{abstract}


\section{Introduction}

With the widespread use of various cameras and the development of three-dimensional computer vision, simultaneous localization and mapping (SLAM) and structure-from-motion (SfM) systems have been extensively studied in the past few decades. In order to improve the accuracy of mapping and localization, it is an indispensable part to optimize the 3D pose information using a back-end algorithm.

\begin{figure}[tp] 
	\centering 
	\subfigbottomskip=2pt 
	\subfigcapskip=-5pt
	\subfigure[KITTI 00]{
		\includegraphics[width=0.47\linewidth]{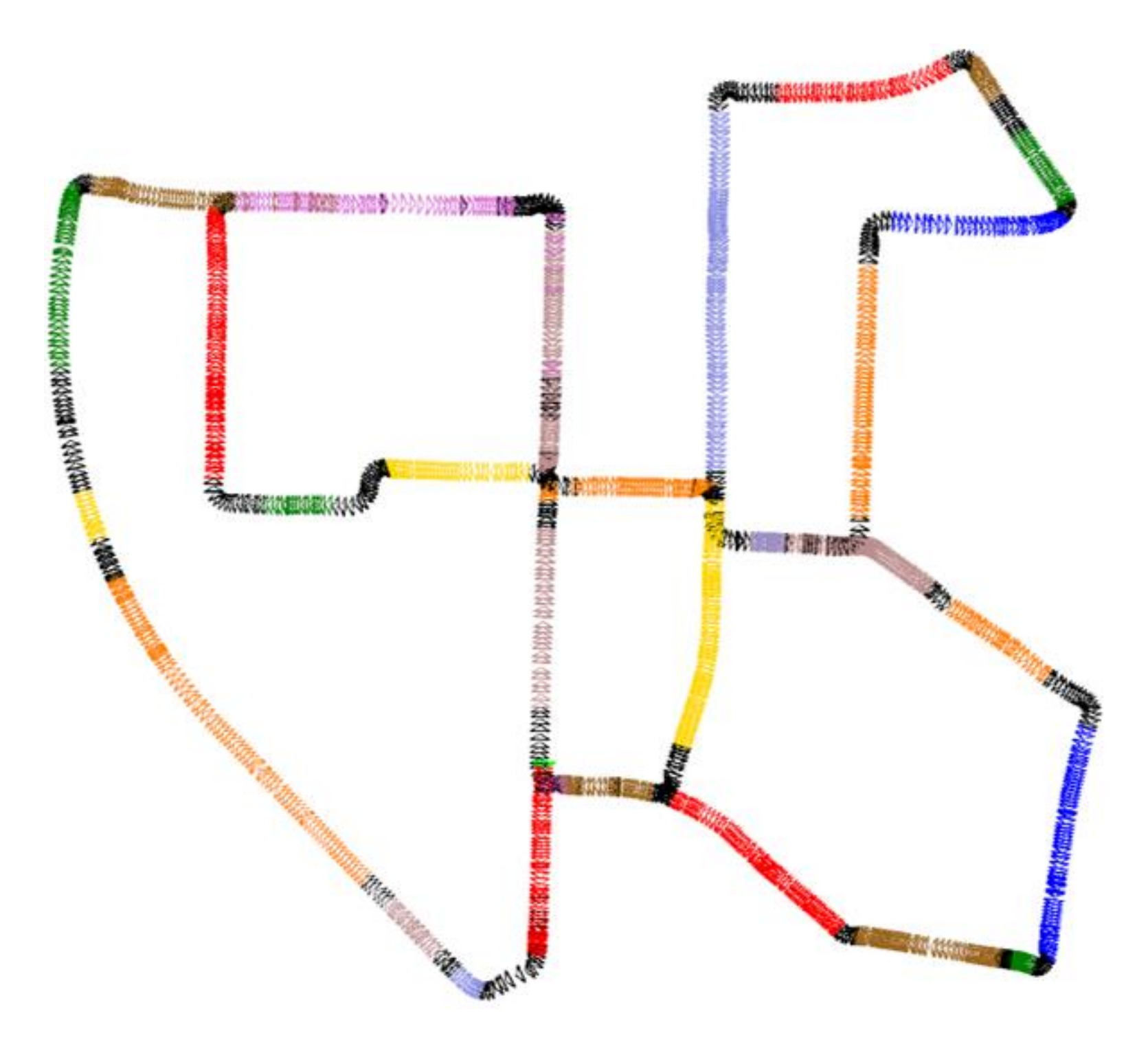}}
	\subfigure[KITTI 05]{
		\includegraphics[width=0.47\linewidth]{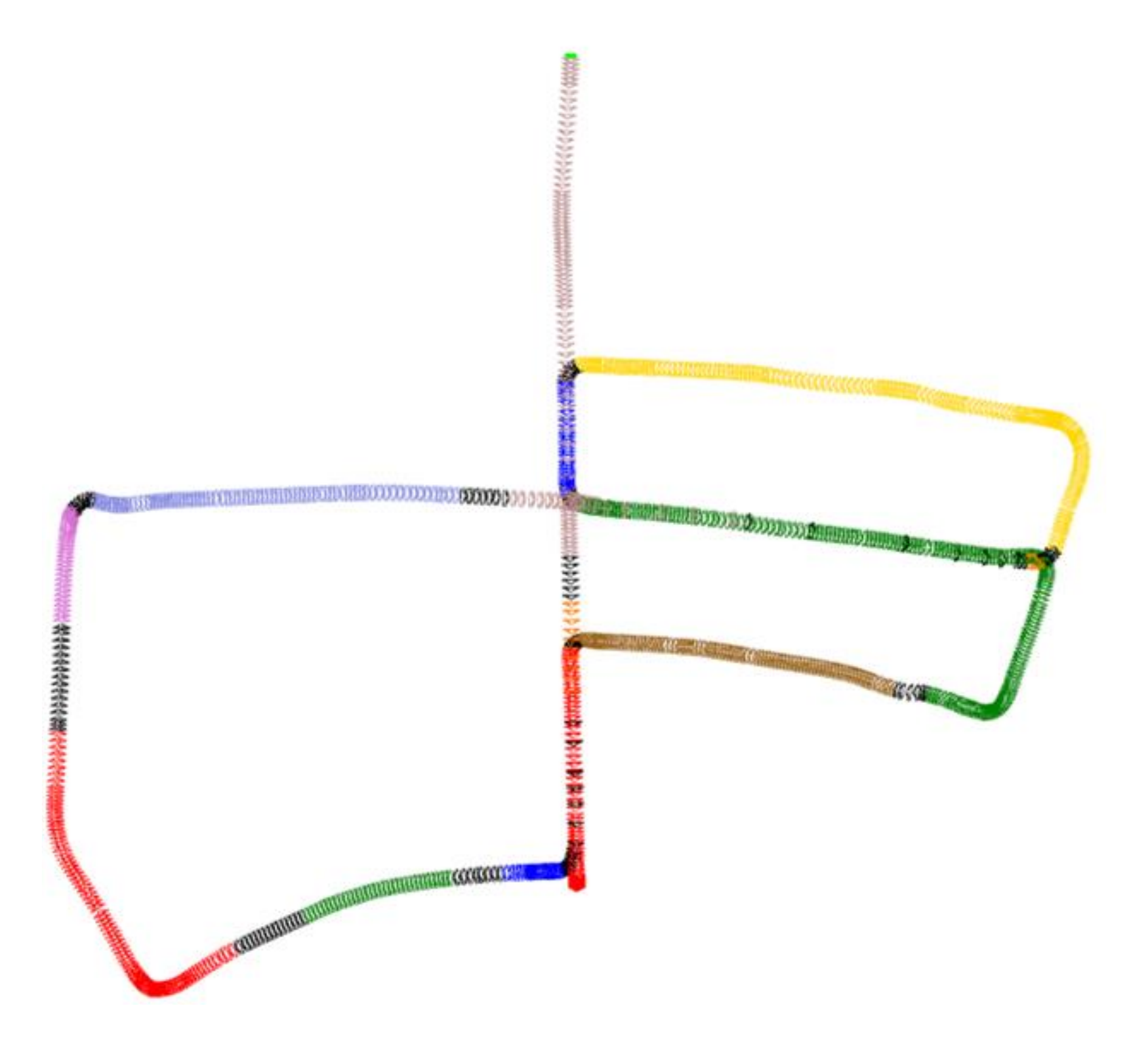}}
	\subfigure[comparison of accuracy and efficiency]{
	    \label{fig:result chart}
		\includegraphics[width=0.97\linewidth]{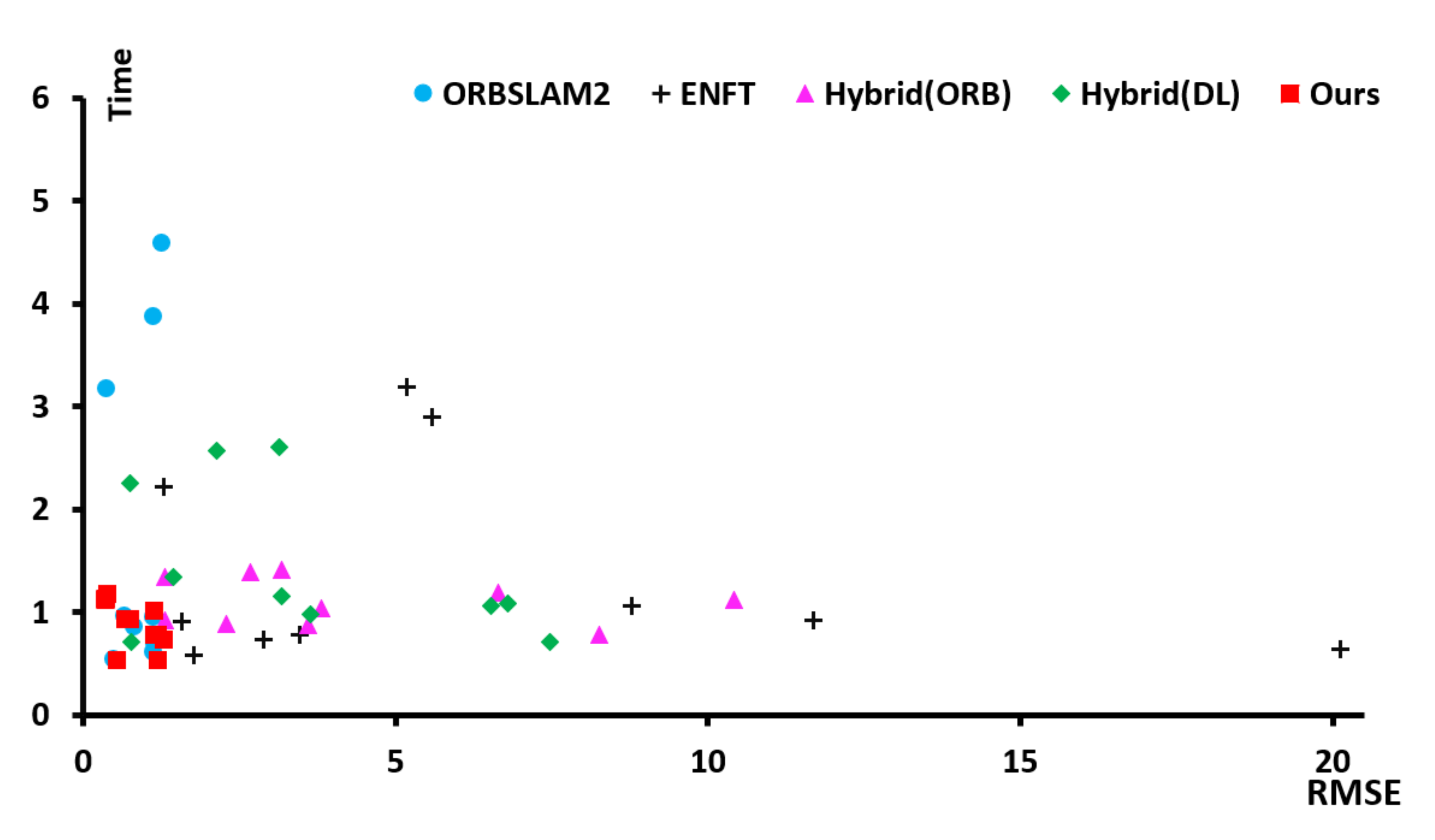}}
	\caption{In the upper part, the figure shows the segmentation result of sequence 00 and 05 of KITTI respectively. The colored areas represent segments, and the black areas represent the buffer. In the lower part, We show the accuracy (RMSE) and efficiency (time) performance of different high-efficiency optimization algorithms on each of the KITTI sequences (each point represents a sequence). Note that the values are divided by the length of the trajectory.}
	\label{fig:seg_actual}
	\vspace{-1cm}
\end{figure}

Bundle adjustment (BA) is an optimization method often used in SfM and SLAM systems. Meanwhile, pose graph is another optimization method that is commonly used in SLAM systems. Unlike BA that optimizes both poses and landmarks (landmark represents the 3D position of the feature point in the world coordinate), pose graph only optimizes poses. Besides, BA uses reprojection error as the optimization target, while pose graph uses relative pose error.

However, the traditional BA and pose graph require excessive computation and usually a large number of iterations, which consumes a lot of computation resources and time. As a result, many real-time, large-scale, and complex multi-robot applications are hindered.

In order to improve the efficiency of the system, many papers propose various improvements to the BA method, while seldom consider that for pose graph. Also, compared with SfM, there are relatively fewer works that aim at improving the efficiency of a SLAM system while maintaining the accuracy of pose estimation. 

So far, there are mainly two categories of methods that can improve efficiency for BA, which are incremental BA and hierarchical BA. Incremental BA mainly follows the idea of augmenting current map by incrementally merging new measurements to reduce the amount of calculation. But this method mainly aims at improving the efficiency of front-end local optimization in the SLAM system and the estimated poses tend to drift due to the lack of global optimization. 

Hierarchical BA often uses a method of dividing the entire trajectory into several segments. Then it uses different methods for optimization within each segment and between segments to improve computation efficiency. Many papers study hierarchical BA from various aspects. However, the current methods have some drawbacks. Some methods\cite{ni2007out, fang2019merge, wu2011multicore} reduce the number of iterations without reducing the complexity of computation, resulting in limited efficiency improvement. Other methods \cite{eriksson2016consensus, li2020hybrid, zhou2020stochastic, zhang2016efficient, hierarchy2014} use local optimization method within each segment and alignment between different segments to replace global optimization, but the accuracy of the system is not ideal due to the lack of error distribution globally. There is also method \cite{zhang2016efficient} that mainly focuses on the error of the connection parts between different segments and omits the optimization within the segment. But this method has high requirement for the accuracy of the segmentation, and it is easy to occur that the connection between the segments is not smooth.

Our work follows the line of hierarchical BA and extends the idea to pose graph optimization. In this paper, we propose a segment-based optimization method suitable for SLAM. Unlike traditional efficient optimization methods that mainly focus on SfM, we make full use of the orderly and rich prior knowledge of SLAM system to improve the efficiency of optimization algorithm. We use the velocity information and the reprojection error information generated during tracking in the SLAM system for segmentation. This method tries to provide a more accurate result for the pose within the segment and a less ideal result for the pose between segments. In this way, when global optimization is performed, more attention will be paid to the areas with poor accuracy, so that we can obtain better optimization results while improving efficiency.

At the same time, we set up a buffer mechanism when segment. we add multiple separate frames as buffer areas between two adjacent segments. For example, Fig. \ref{fig:seg_actual} shows examples of the segmentation result. We optimize all the frames inside the buffer areas when performing global optimization. This method makes full use of global information to reduce the errors between different segments to achieve higher accuracy results. Besides, the use of the buffer area can also alleviate the problem of segmentation accuracy. In this way, we no longer need to pay much attention to the exact position of splitting point, but rather only an approximate position, which can effectively improve robustness.

Finally, when performing global  optimization, the first and last several frames of each segment and the frames in the buffer areas are globally optimized. Then those frames that have participated in the global optimization phase in each segment are used for interpolation to update pose information within the segment. The specific algorithm is described in detail at Section III. We apply the method to public SLAM datasets including KITTI \cite{KITTI}, TUM RGB-D \cite{tum-rgbd} and EuRoC \cite{euroc}, and the results show that our approach significantly outperforms existing methods in terms of efficiency and accuracy, as shown in Fig. \ref{fig:result chart}.

In summary, the contributions of this paper include:
\begin{itemize}

\item We propose a hybrid trajectory segmentation and buffering method that can be used in the back-end optimization of the SLAM system to improve the efficiency of computation, meanwhile alleviating the problem of excessively high requirements for segmentation accuracy.

\item We are the first to introduce interpolation-based method to replace heavy optimization to efficiently calculate the inner-segment poses, while ensuring that there is almost no loss in accuracy.  

\item We propose a unified segment-based optimization method for both pose graph and global BA in SLAM, and our system significantly outperforms the previous method in both efficiency and accuracy.



\end{itemize}

\begin{figure*}[tp] 
	\centering
	\includegraphics[width=1\linewidth]{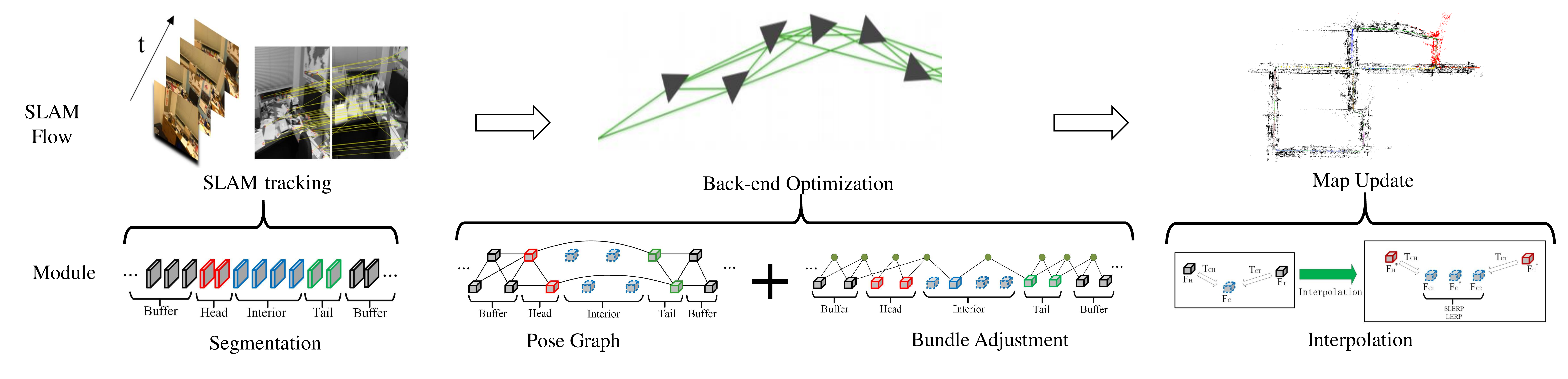}
    \caption{\textbf{Pipeline of proposed method.} The upper part shows the overall process of existing SLAM system, and the lower part shows the main ideas of our segment-based method for each corresponding process.}
    \label{fig:flow}
    \vspace{-0.5cm}
\end{figure*}

\section{Related Works}

There are various kinds of approaches aiming at solving the efficiency problem caused by the increasing number of frames in the phase of optimization such as bundle adjustment, especially in large-scale environments. Early work \cite{engels2006bundle} leverages the sparsity pattern of Hessian matrix in BA and proposed to use Schur complement to compute camera poses first and then use back substitution for landmarks. It significantly reduces the complexity of BA and is later introduced into common solvers like g2o \cite{kummerle2011g}. 

One line of research \cite{kaess2008isam, liu2018ice, ila2017fast, indelman2012incremental, vo2016spatiotemporal, schonberger2016structure, shah2015multistage} uses the method of incremental BA, where the new frames are incrementally merged to the existing map. However, the performance of the system suffers from poor initialization and accumulated error, especially for long sequences.

Besides, there is a growing trend of hierarchical BA, which divides the optimization into several phases and in each phase only sub-group of nodes are optimized \cite{suger2014approach, ni2007out, fang2019merge, wu2011multicore, eriksson2016consensus, li2020hybrid, zhou2020stochastic, zhang2016efficient}. To select the sub-group of optimization nodes, the most commonly used method is to partition the trajectory into several segments, where the error within each segment is small and error between segments is large. In this way, we can start by solving small sub-problems that are much easier and faster to convergence, and then design different segment merging strategies to bind the segments together. As a result, the overall efficiency of BA can be improved and the optimization result can be more stable. In \cite{ni2007out}, images are first clustered based on similarity metric and local BA is carried out to optimize poses within the cluster, and then global BA is used to merge clusters. \cite{fang2019merge} is similar in concept except for binding segments using pairwise alignment. \cite{wu2011multicore, eriksson2016consensus} tries to compute in parallel for clusters, thus better enhancing efficiency. \cite{zhou2020stochastic, zhang2016efficient} changes the way of partitioning in an adaptive way between iterations during optimization, which introduces extra complexity cost. But these methods mainly aim at solving the offline SfM problem.

For segment partitioning criterion, co-visibility \cite{li2020hybrid,zhou2020stochastic} are often used as the criterion. For example, \cite{li2020hybrid} designs a hybrid method for SLAM system, which leverages both spatial and temporal co-visibility to determine the connectivity of the current frame with nearby frames and splits the trajectory accordingly. However, it is doubtful that co-visibility can reliably reflect the error metric. For example, at the place where co-visibility is large, the error is not necessarily small, since we cannot guarantee the accuracy of the 3D location of landmarks. \cite{zhang2016efficient} uses the angle of steepest descent direction between consecutive frames as the criterion to partition the trajectory, which only consider local variations. Besides, \cite{zhang2016efficient} claims that although reprojection error serves as a solution, it may break the trajectory into too many segments since the large error tends to distribute in a region where the splitting point is chosen. However, we argue that this problem can be solved by introducing buffer regions between segments, and all the frames within the buffer would be treated as the large-error region that require subsequent optimization. Along with reprojection error, we also use velocity information to better identify the splitting point and the system is able to run reliably.

To merge the segments together, it is nature to use frames at the edge of each segment for alignment or optimization. For example, \cite{li2020hybrid} uses local BA to optimize pose within segment and uses alignment method based on motion averaging to calculate the relative transformation between segments and merge them together. Since only pairwise transformations are estimated, it can easily cause accumulated error. Therefore, while improving efficiency of the BA process, the accuracy of the result is significantly impacted. Instead, \cite{zhang2016efficient} proposed to treat each segment as a rigid body by assigning a single 7-DoF similarity transformation, and applied global BA to optimize only the overlapping frames, thus avoiding drift error. The poses of the frames within the segment are updated by relative pose change before and after optimization. However, the alignment is easy to cause inconsistency between segments. To solve this, the partitioning and optimization process is carried out hierarchically, which is tedious. Moreover, we argue that assigning two 7-DoF similarity transformations at both the head and tail part of the segment can easily solve the inconsistency problem. The pose of frames within the segment can be obtained by interpolation method similar to \cite{jang2020pose} and the accuracy of the system remains almost the same.

\section{Proposed Method}
\label{method}
In the current SLAM framework, the optimization of the back-end mainly has two parts, one is the pose graph optimization using the camera pose information, and the other is the bundle adjustment optimization using the reprojection error. The segmentation method we propose can effectively improve computational efficiency in both optimization modules. Specifically, we divide our method into the following five parts for detailed description.

Section~\ref{segmentation method} shows how we use the prior information of the SLAM system for segmentation. Section~\ref{buffer mechanism} describes the proposed buffer mechanism in detail. Section~\ref{Pose Graph Optimize} and Section~\ref{global bundle adjustment optimize} describes the way to incorporate proposed method into pose graph optimization and bundle adjustment optimization module respectively. Finally, Section~\ref{Interpolation within segment} describes interpolation method for the camera pose within the segment after global optimization. The pipeline of our method is shown in Fig. \ref{fig:flow}.


\subsection{Segmentation Method} 
\label{segmentation method}
To improve the efficiency of the system, the global optimization should mainly optimize the parts with lower accuracy, and pay less attention to the parts with better accuracy. Therefore, the ideal frame segmentation should follow: 
\begin{itemize}
\item For each segment, the pose estimation of the frames and the landmark within segment is relatively accurate.
\item The pose estimation at the connection parts between the segments have a relatively large error. 
\end{itemize}


In order to achieve the goal, we analysis the mathematical form of pose graph optimization \eqref{form1} and bundle adjustment optimization \eqref{form2} in the SLAM system. 
\begin{equation}
\label{form1}
min\sum_{{i,j}\in\varepsilon}{{e_{ij}}^{T}\Sigma_{ij}^{-1}e_{ij}}
\end{equation}
where
$$
e_{ij}={ln(\Delta{T_{ij}}^{-1}{T_i}^{-1}T_j)}^\vee
$$
\begin{equation}
\label{form2}
min\sum_{i=1}^{n}\sum_{j=1}^{m}\Sigma_{ij}^{-1}{(u_{ij}-\pi(L_j,F_i))}
\end{equation}
In equation \eqref{form1}, $T_{i}$ means the pose of the $i$-th node and $T_{ij}$ means the relative pose between $i$-th node and $j$-th node. The symbol $^\vee$ transform a matrix to vector, thus $e_{ij}$ represents the error vector. $\Sigma_{ij}$ represents the covariance of the edge. In equation \eqref{form2}, $u_{ij}$ and $L_j$ represent the 2D coordinates of the j-th feature points on the i-th frame and 3D position of the corresponding landmark. $F_i$ means the pose of i-th frame and $\pi$ means the projection of the landmark from 3D to 2D. 


The best splitting points in the graph optimization should be near the edges with larger error. According to \eqref{form1}, the segmentation points in the graph should be the frames with poor estimation of the adjacent pose transformation in the pose graph optimization. And according to \eqref{form2}, the segment points in the optimization graph should be frames with large reprojection errors in the bundle adjustment optimization. Therefore, the splitting point should be the frames with large reprojection error or large relative pose error.

During the tracking process of the SLAM system, estimation errors will inevitably occur due to the measurement noise, but these errors are generally small. However, some drastic changes in the environment can cause a sudden increase in pose estimation errors, such as collisions, drifts, or feature point detection and matching errors. These conditions are reflected in the SLAM system as rapid changes in speed or increased reprojection errors. The location of the splitting point should reflect these unexpected changes.

Therefore, we use the pose transformation speed of adjacent frames and the reprojection error of each frame as the criterion for segmentation. Equation \eqref{form3} shows the mathematical form of the criterion, where $F$ represent frame and $F_C$ means the current frame. $\vec{v}$ means the pose transformation velocity vector and $reproj$ represents the reprojection error of the landmarks observed by the frame. $\varphi$ means the set of all frames in the segment and $\delta(F_C)$ means the set of landmarks that can be observed by the current frame. In addition, $\sigma_v$ and $\sigma_r$ are threshold parameters. If the following two conditions hold, the current frame will be categorized as part of the segment. If it does not meet one of the conditions, the current segment ends and it will be categorized as the buffer area. And the next segment will begin according to Equation \eqref{form4}. Details about  buffer area and Equation \eqref{form4} will be discussed in Section~\ref{buffer mechanism}.
\begin{equation}
\label{form3}
\left\{\begin{array}{lc}||\vec{v}(F_C)-\frac{1}{m}\sum_{F_i\in \varphi}{\vec{v}(F_i)}||&<\sigma_v \\ reproj(F_C)&<\sigma_r
\end{array}\right.
\end{equation}
where 
$$
reproj(F_i)=\frac{1}{|\delta(F_C)|}\sum_{L_j\in\delta(F_C)}{||u_j-\pi(L_j,F_i)||}
$$

Compared with the previous segmentation methods \cite{fang2019merge,li2020hybrid,zhou2020stochastic} which mainly use the number of common view landmarks as the segmentation criterion, our approach is more closely related to the optimization equation, more quantitative, and more relevant to the final positioning accuracy. The ablation study of different segmentation methods can be found in Table \ref{table:Segment}.

\subsection{Buffer Mechanism}
\label{buffer mechanism}

 \begin{figure}[tp] 
	\centering
	\subfigbottomskip=2pt 
	\subfigcapskip=-5pt
	\subfigure[no buffer]{
		\label{fig:no buffer}
		\includegraphics[width=0.97\linewidth]{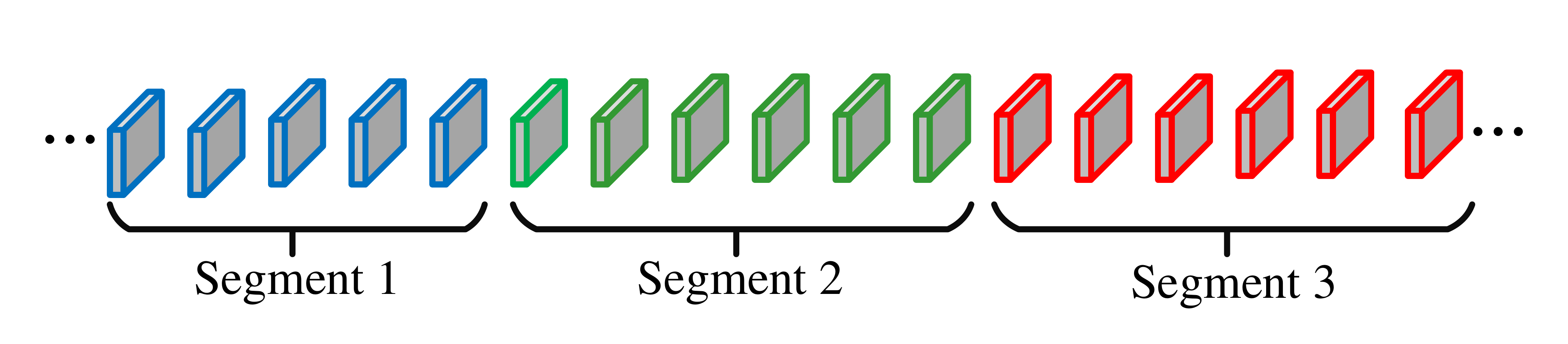}}
	\subfigure[buffer]{
		\label{fig:buffer}
		\includegraphics[width=0.97\linewidth]{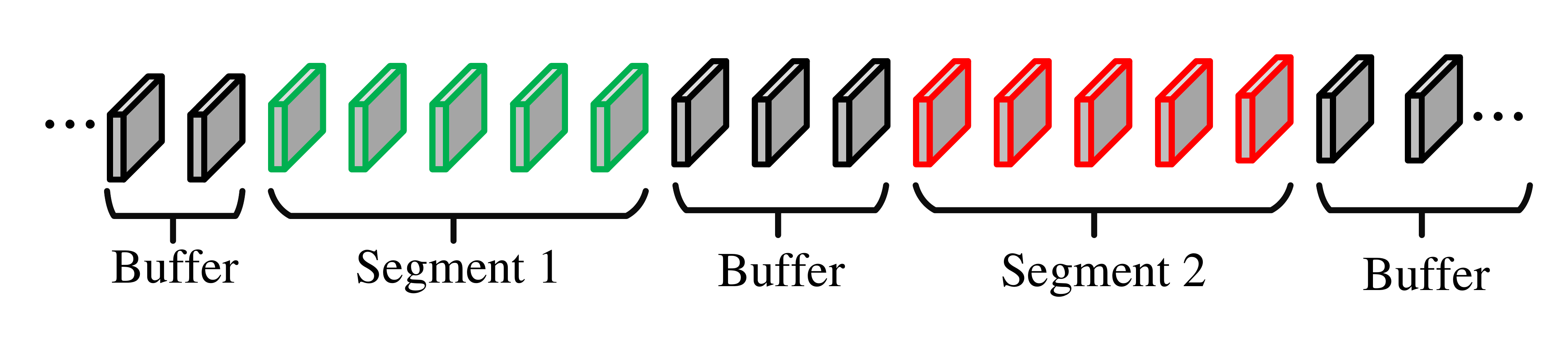}}
	\caption{\textbf{Illustration of buffer mechanism}. The segments are composed of colored frames and the buffers are composed of black frames.}
	\label{fig:buffer result}
	\vspace{-0.7cm}
\end{figure}

For the traditional segmentation method, two adjacent segments are directly connected, as shown in Fig. \ref{fig:no buffer}. Therefore, the specific location of the splitting point needs to be handled carefully when segmenting, because different segmentation can yield different output, which can affect the robustness of the optimizer. In addition, it is commonly seen that the pose estimation of multiple consecutive frames are all not accurate in the SLAM system. It is unreasonable to regard these frames as a segment.

In order to solve this problem, we design a buffer mechanism. Between the divided two segments, a buffer is placed, as shown in Fig.\ref{fig:buffer}. In the process of segmentation, first use the method in section A to determine whether the current section is going to end. Then put the following frames into the buffer until both the speed change and re-projection error have recovered to a relatively stable level. Finally, create a new segment and repeat the above process. And the criterion for the end of selection for the buffer area is shown in \eqref{form4}. $\eta_v$ and $\eta_r$ represents the stability of velocity and reprojection error. $\alpha$ and $\beta$ are weight factors. When $\eta_v(F_i)$ is less than a threshold, a new segment will be created. For experiments, we define $\alpha$ as 0.2, $\beta$ as 0.8 and threhold as 0.5, and we can achieve the desirable performance. 

\begin{figure}[tp] 
	\centering
	\includegraphics[width=0.97\linewidth]{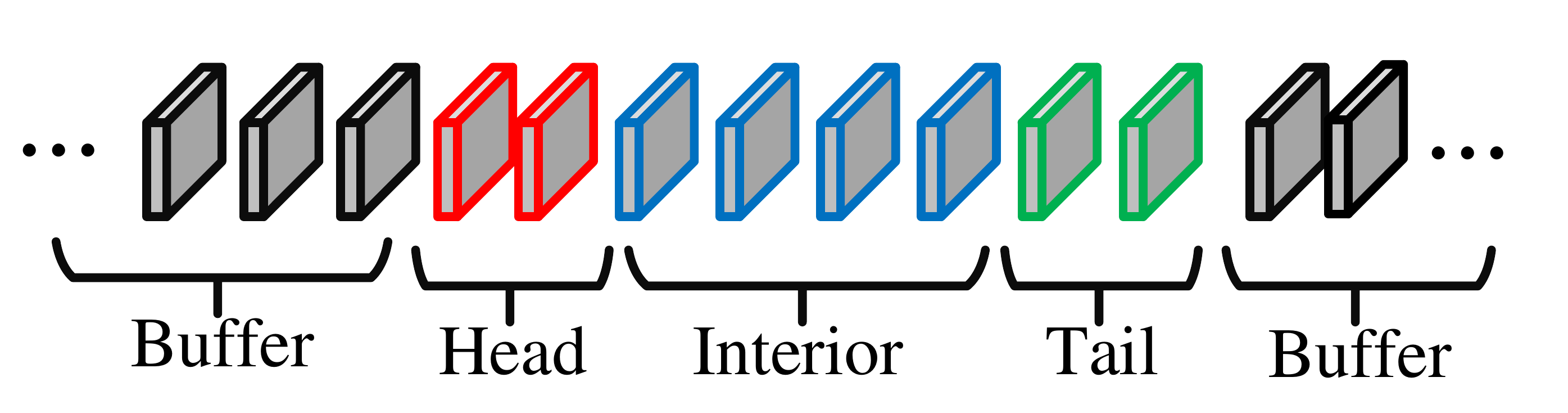}
    \caption{\textbf{Definition for segmentation result}. Red, blue and green frames respectively represent head, interior and tail part of the segment. Each segment consists of these three parts. And frames in the buffer area are shown in black.}
    \label{fig:seg_res}
    \vspace{-0.5cm}
\end{figure}

\begin{equation}
\label{form4}
\eta(F_i)=\alpha*\eta_v(F_i)+\beta*\eta_r(F_i)
\end{equation}
where 
$$
\eta_v(F_i) = \frac{||\vec v(F_C)-\frac{1}{|\varphi|}\sum_{F_i\in \varphi}{\vec {v}(F_i)}||}{||\frac{1}{m}\sum_{F_i\in \varphi}{\vec {v}(F_i)}||}
$$
$$
\eta_r(F_i) = \frac{||reproj(F_C)|-\frac{1}{|\varphi|}\sum_{F_i\in \varphi}{|rerpoj(F_i)|}|}{|\sum_{F_i\in \varphi}{|reproj(F_i)|}|}
$$

At the same time, in order to facilitate subsequent optimization, we divide each segment into three parts: head, tail, and inside according to the order of the frames in the segment. In experiments, we use the first two frames as the head part and the last two frames as the tail part. In the end, each frame in the SLAM system belongs to one of the four categories: head, tail, inside, and buffer. The definition of segmentation result is shown in Fig. \ref{fig:seg_res}.

The buffer mechanism effectively alleviates the excessive requirements of segmentation accuracy in traditional segmentation methods, and improves the robustness of the optimizer. The design of buffer mechanism can also flexibly handle continuous inaccurate pose estimation. At the same time, the use of the buffer area hardly produces extra computational cost for the system. Compared with the traditional segmentation method, it sacrifices very little efficiency in exchange for the improvement of robustness and accuracy.

\subsection{Pose Graph Optimization}
\label{Pose Graph Optimize}
\begin{figure}[tp] 
	\centering
	\subfigbottomskip=2pt 
	\subfigcapskip=-5pt
	\subfigure[traditional]{
		\label{fig:pose graph ori}
		\includegraphics[width=0.97\linewidth]{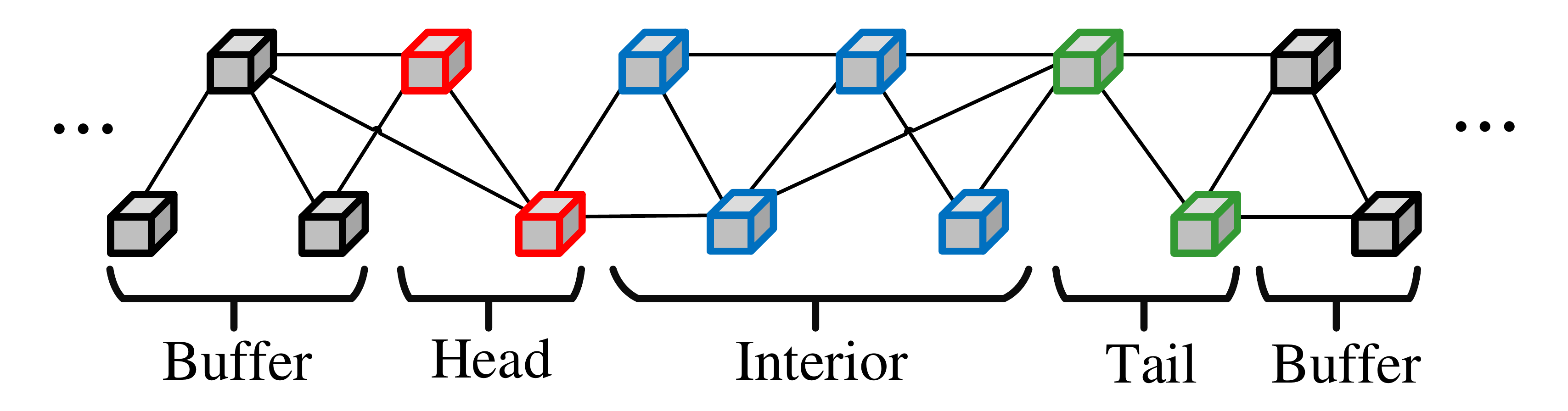}}

	\subfigure[our method]{
		\label{fig:pose graph new}
		\includegraphics[width=0.97\linewidth]{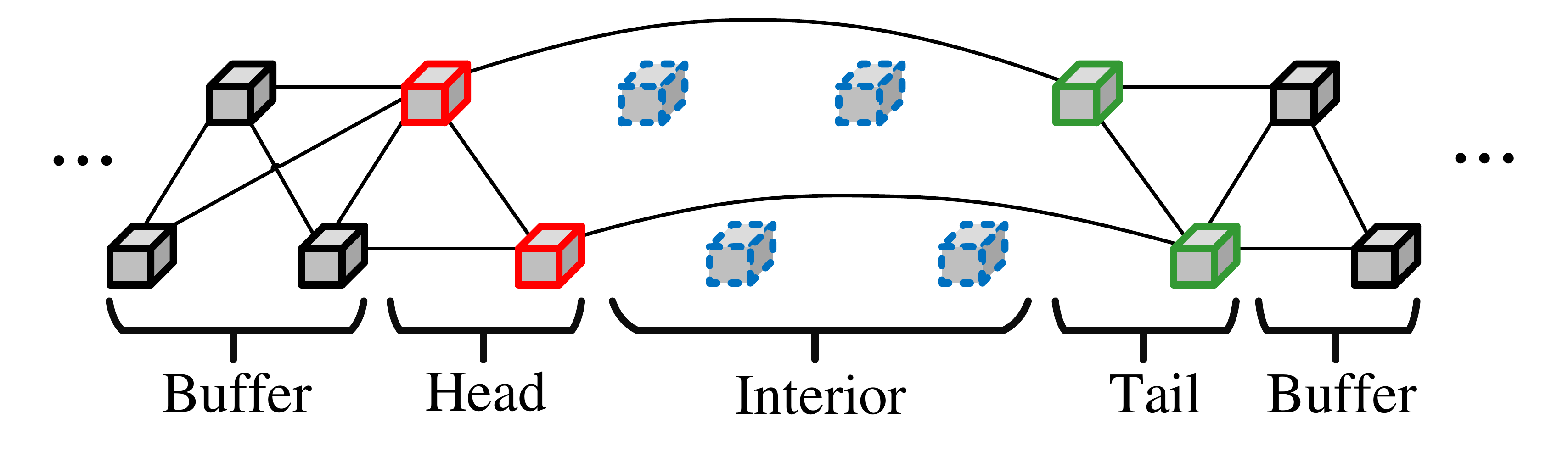}}
	\caption{\textbf{Pose Graph Method.} The solid-line squares represent the pose nodes in the optimization graph. The dotted-line squares represent the pose nodes that require updating via interpolation. And the connection line represents the relative pose transformation that is used as edge during optimization.}
	\label{pose graph method}
	\vspace{-0.5cm}
\end{figure}

The pose graph optimization in the SLAM system mainly uses the relative pose transformation of the frame to optimize. The nodes in the optimization graph represent the poses of the frames in the world coordinate system, and the edges between the nodes represent the relative pose transformation. The optimization graph in traditional pose graph optimization is shown in Fig. \ref{fig:pose graph ori}.

In our method, we change the optimization equation via segmentation information to improve the efficiency of optimization. We divide the entire optimization process into two steps: global optimization and pose update within segment. In the following sections, we follow the definition of the four categories of segmentation shown in Fig. \ref{fig:seg_res}.

In the global optimization stage, we use the segmentation information to reduce the number of nodes and edges in the pose graph, so as to reduce the amount of calculation of the optimizer. We delete all nodes in the interior part of segment and their related edges. But in order to ensure the connectivity of the pose graph during global optimization, we connect edges between nodes in the head of segment and nodes in the tail  to approximate the deleted edges. Finally, the nodes and edges diagram of the pose graph that needs to be optimized is shown in Fig. \ref{fig:pose graph new}. As the nodes and edges in the pose graph have changed, the optimization equation also changes, as shown in \eqref{form5}, where $\varphi(S)$ represents the set of frames in the interior part of segment. By reducing the number of nodes and edges participated in optimization, we greatly improve the speed of optimization.
\begin{equation}
\label{form5}
min\sum_{{\tiny \begin{array}{c}
\{i,j\}\in \Omega \\ i,j\notin
\varphi(S)\end{array}}}{{e_{ij}}^{-1}\Sigma_{ij}^{-1}e_{ij}}
\end{equation}

After pose graph optimization, it enters the stage of pose update within the segment. From the previous step, we can only get the optimized result of nodes in head, tail and buffer, thus we need to use another method to get the updated result for frames in the interior part of segment. For each segment, we fix the frames in the head and tail, and use more efficient interpolation instead of optimization to update the nodes in the interior. The interpolation method will be described in detail in section \ref{Interpolation within segment}.  In this way, the pose of each frame in the SLAM system has the final optimized result.

\begin{figure}[tp] 
	\centering
	\subfigbottomskip=2pt 
	\subfigcapskip=-5pt
	\subfigure[traditional]{
		\label{fig:ba ori}
		\includegraphics[width=0.97\linewidth]{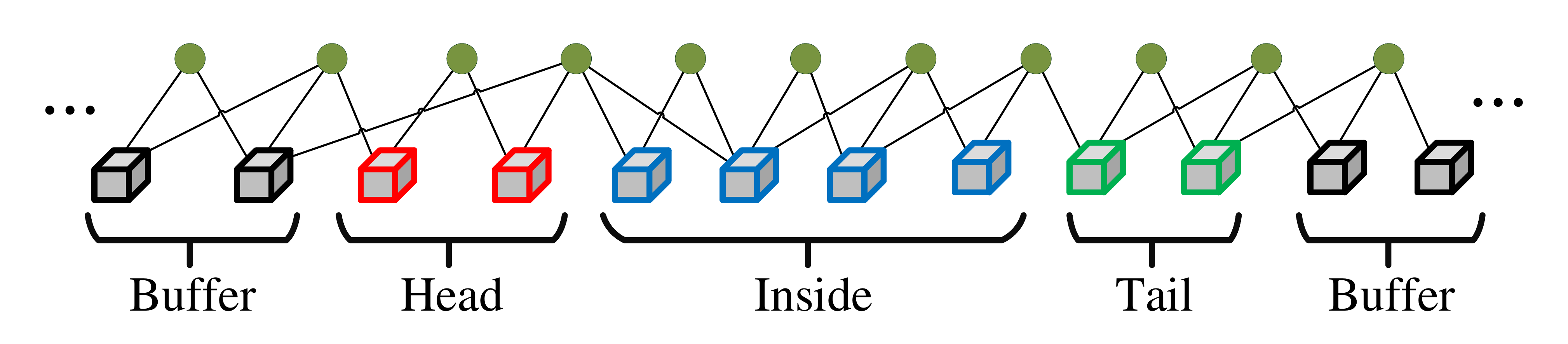}}

	\subfigure[our method]{
		\label{fig:ba new}
		\includegraphics[width=0.97\linewidth]{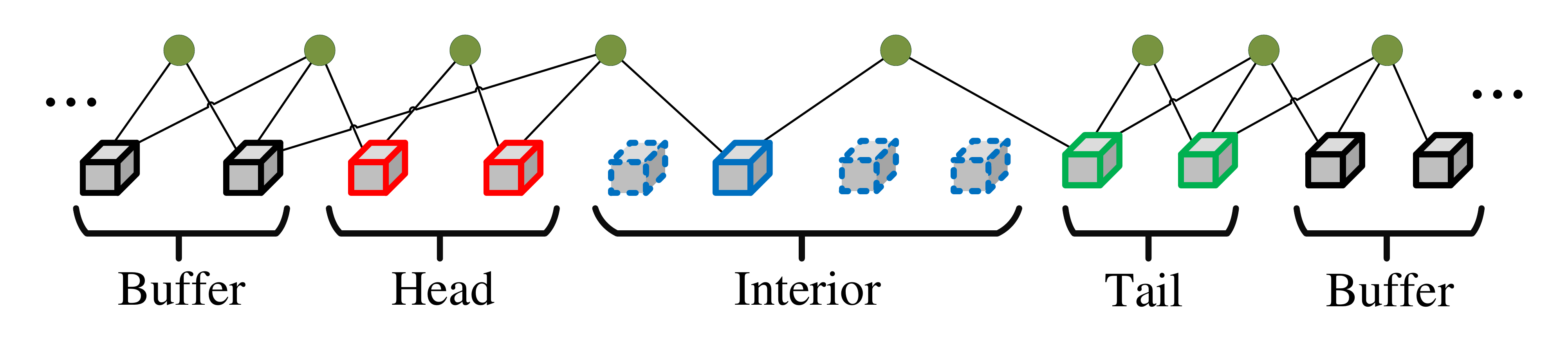}}
	\caption{\textbf{Bundle Adjustment Method.} The solid-line squares represent the frame nodes in the optimization graph and the dotted-line squares represent the pose nodes that require updating via interpolation. In particular, the solid-line squares in the interior part of segment represent connecting frames. The green dots represent the landmark nodes and the connection lines represent the reprojection-error edges.  }
	\label{ba method}
	\vspace{-0.5cm}
\end{figure}

\subsection{Bundle Adjustment Optimization}
\label{global bundle adjustment optimize}
The bundle adjustment optimization in the SLAM system uses the reprojection error between landmarks position and frame pose to optimize. The nodes in the optimization graph represent the pose of the frames and the 3D coordinates of the landmarks, and the edges represent the estimation of the reprojection error from landmarks to frames.

Similar to the pose graph optimization method, the entire optimization process is also divided into two steps: global pose optimization and pose update within segment.

In the global optimization stage, the nodes and edges are first simplified. Similar to the pose graph method, we keep the frame nodes which are in head, tail and buffer. Besides, in order to ensure the connectivity, we choose the least number of interconnected frames as the connecting frames between the head and the tail of each segment. In practice, we select the last frame in time sequence from the frames whose covisibility with the last frame in the head area are greater than 30 and add it to the connecting frames. Then use the same method for the the frame just selected to obtain the new connecting frame. Repeat the above process until the covisibility of the last selected frame with the first frame in the tail area exceeds 30. In this way, we choose the fewest connecting frames to maintain the connectivity of the frames for the bundle adjustment. In the end, the frame nodes that we execute global optimization includes frames which are in head, tail, buffer and the connecting frames. For the landmark nodes, only nodes that have connected edges to the frame node are retained. The origin optimization graph for BA is shown in Fig. \ref{fig:ba ori} and our proposed one is shown in Fig. \ref{fig:ba new}. As the nodes and edges in the optimization graph change, the optimization equation also changes which is shown in \eqref{form6}, where $\varphi(H)$ and $\varphi(T)$ represents the set of frames in the head and tail parts of segment respectively. And $\varphi(C)$ represents the set of connected frames.
\begin{equation}
\label{form6}
min\sum_{F_i\in \varphi^{\ast}}\sum_{L_j\in C(F_i)}{\Sigma_{ij}^{-1}(u_{ij}-\pi(L_j,F_i))}
\end{equation}
where 
$$\varphi^{\ast} = \varphi(H)\cup \varphi(T)\cup \varphi(C)$$

After global bundle adjustment optimization, it enters the stage of pose update within the segment. Similar to pose graph module, for each segment, we fix the frames in head, tail, and connecting frames. Then use interpolation instead of optimization to update other nodes in the interior part of segments and then align the relative landmark position. In this way, every frame poses and landmark coordinates in the SLAM system will get the final optimized results.

\subsection{Interpolation within segment}
\label{Interpolation within segment}
In this section, we will introduce the interpolation method. We use this method to obtain approximate results for the unoptimized frames based on the optimized frames after the global optimization stage. 

\cite{jang2020pose} shows that when there are accurate pose results obtained by optimization on both sides of the frame which needs to be estimated and accurate estimations of the pose transformation relationship between the frames, we can use interpolation instead of optimization to get approximate optimization results which can avoid heavy computation. Therefore, we use an improved interpolation method instead of the optimization method to efficiently calculate the results.

In pose graph optimization module, frames in the head and tail part of the segment will participate in global optimization and get more accurate pose results. Besides, according to the segmentation strategy in section A, the accuracy of pose transformation between frames in the segment is also relatively high. We use frame poses in head and tail as reference. And for each frame in the interior of segment that doesn't participate in global optimization, we use the transformation relationship between the current frame and the reference results to accomplish interpolation. 
$$
{R_{HC}^\ast}=R_{HC}\cdot SLERP(\delta(R),\alpha)
$$
\begin{equation}
\label{form7}
{t_{HC}^\ast}=t_{HC}+LERP({R_{HC}^\ast}\cdot\delta(t),\alpha)
\end{equation}
$$
s_C={\frac{1}{\alpha+1}s_H}+{\frac{\alpha}{\alpha+1}s_T}
$$

where
$$
\delta(R)={R_{HC}}^T\cdot{R_{HT}}^\ast\cdot R_{TC}
$$
$$
\delta(t)=t_{HC}+{R_{CH}}^\ast t_{HT}+{R_{CT}}^\ast t_{TC}
$$
$$
\alpha=\frac{\sqrt{\sum_{F_i\in\varphi(l)}{||\vec{v}(F_i)||}^2}}{\sqrt{\sum_{F_i\in\varphi(r)}{||\vec{v}(F_i)||}^2}}
$$

The mathematical form of the interpolation algorithm is shown in \eqref{form7}, where $R$, $t$ and $s$ represent rotation matrix, translation vector and scale. The subscripts $C$, $H$ and $T$ indicate current frame, reference frames used for interpolation that are before and after the current frame. $\varphi(l)$ and $\varphi(r)$ represent the set of frames in a similar way. $SLERP$ means quaternion spherical interpolation and $LERP$ means linear interpolation.

We use relative velocity as the interpolation factor. Then we use relative pose transformation to calculate the rotation matrix of the current frame based on the results in the head and tail respectively. Finally, spherical linear interpolation is used to obtain the final rotation result. And the interpolation result of translation is also calculated using a similar method.

For bundle adjustment module, connecting frames along with frames in the head and tail part of the segment will all participate in global optimization. Therefore, the reference frames for interpolation will also include connected frames in addition to frames in the head and tail parts of segments. And the rest are the same as that in the pose graph module. After completing the interpolation for the pose of the frame, the landmarks that did not participate in the bundle adjustment will be updated according to their positional relationship with the reference frame.

Due to the high accuracy of estimation from global optimization, the use of interpolation instead of optimization can obtain results in a small loss of accuracy but can save a lot of time. At the same time, compared with the method of directly not optimizing the error within segment such as \cite{zhang2016efficient}, interpolation can effectively eliminate the error. Our method can strike a better balance between efficiency and accuracy.

\section{Evaluation}
\begin{table}[tp]
\begin{center}
\caption{\textbf{Base Line Experiment.} We selected sequences with loops in three SLAM datasets for this experiment. We record the time of pose graph optimization and bundle adjustment optimization during each loop to evaluate efficiency, and record the final RMSE of each sequence to evaluate accuracy. PG means the time of pose graph optimization in each loop and BA means the time of bundle adjustment. The unit of PG and BA is second(s). The unit of RMSE is meter(m) in KITTI sequences and centimeter(cm) in TUM and EuRoC sequences.}
\label{table:Base Line}
\setlength{\tabcolsep}{1mm}
\begin{tabular}{c|c|ccc|ccc}
\hline
\multicolumn{2}{c|}{} & \multicolumn{3}{c|}{\tabincell{c}{OpenVSLAM \\{[8]}}} & \multicolumn{3}{c}{\tabincell{c}{Our\\method}}\\
\cline{3-8}
\multicolumn{2}{c|}{}& PG & BA & RMSE & PG & BA & RMSE\\
\hline

\hline
\multirow{4}*{\tabincell{c}{KITTI\\00}} & Loop-1 & 2.01 & 4.68 & \multirow{4}*{1.32}& 0.50 & 1.49 &\multirow{4}*{1.33}\\
\cline{2-4}\cline{6-7}
&Loop-2 & 3.38 & 7.77 & & 0.68 & 2.69 &\\
\cline{2-4}\cline{6-7}
&Loop-3 & 4.24 & 9.67 & & 0.97 & 3.12&\\
\cline{2-4}\cline{6-7}
&Loop-4 & 5.48 & 12.7 & & 1.48& 3.77&\\
\hline

\hline
\multirow{2}*{\tabincell{c}{KITTI\\02}} & Loop-1 & 4.67 & 12.8 & \multirow{2}*{5.71}& 1.67 & 3.87 &\multirow{2}*{5.72}\\
\cline{2-4}\cline{6-7}
&Loop-2 & 4.72 & 13.6 & & 1.87& 3.92\\
\hline

\hline
\multirow{3}*{\tabincell{c}{KITTI\\05}} & Loop-1 & 1.86 & 5.18 & \multirow{3}*{0.83}& 0.23 & 0.78 &\multirow{3}*{0.85}\\
\cline{2-4}\cline{6-7}
&Loop-2 & 3.79 & 8.67 & & 0.53 & 1.77 &\\
\cline{2-4}\cline{6-7}
&Loop-3 & 3.88 & 8.98 & & 0.53 & 1.87 &\\
\hline

\hline
\tabincell{c}{KITTI\\06} & Loop-1 & 1.30 & 5.04 & 0.83 & 0.23 & 0.75 & 0.82\\
\hline

\hline
\tabincell{c}{KITTI\\07} & Loop-1 & 1.89 & 2.66 & 0.49 & 0.18 & 0.82 & 0.52\\
\hline

\hline
\tabincell{c}{KITTI\\09} & Loop-1 & 1.96 & 4.42 & 1.97 & 0.43 & 1.42 & 2.04\\
\hline

\hline
\tabincell{c}{EuRoC\\MH-05} & Loop-1 & 1.25 & 3.48 & 5.76 & 0.26 & 0.46 & 5.83\\
\hline

\hline
\tabincell{c}{EuRoC\\V1-02} & Loop-1 & 0.17 & 0.56 & 6.46 & 0.11 & 0.56 & 6.52\\
\hline

\hline
\tabincell{c}{EuRoC\\V1-03} & Loop-1 & 0.05 & 0.19 & 6.42 & 0.03 & 0.11 & 6.50\\
\hline

\hline
\tabincell{c}{TUM\\loop} & Loop-1 & 4.01 & 13.1 & 9.62 & 1.69 & 2.41 & 9.71\\
\hline

\hline
\tabincell{c}{TUM\\pioneer} & Loop-1 & 1.64 & 3.05 & 6.74 & 0.42 & 2.57 & 6.87\\
\hline

\hline
\multicolumn{2}{c|}{Avg change rate} & +0\% & +0\% & +0\% & -72.2\% & -65.0\% & +1.60\%\\
\hline

\hline
\end{tabular}
\end{center}
\vspace{-0.5cm}
\end{table}
In this section, we conduct experiments of different optimization algorithms, mainly comparing the efficiency and accuracy of the methods. The segmentation method proposed in this article is mainly applicable to SLAM systems. Therefore, our experiments are also carried out in SLAM datasets including KITTI \cite{KITTI}, TUM RGB-D \cite{tum-rgbd} and EuRoC \cite{euroc}.

We have implemented our optimization algorithm based on OpenVSLAM, an open-source SLAM system. So first we compare our optimization method with the traditional method used in OpenVSLAM\cite{sumikura2019openvslam} to verify our improvement. In addition, we also compare our algorithm with other high-efficiency optimization algorithms to reflect the performance of our method in terms of efficiency and accuracy. Finally, we conduct ablation experiments to verify the effects of each module in our method. 

\subsection{Baseline Comparison}
In the comparison experiment with the baseline, we choose the sequences with loops in the SLAM datasets for experiment. We separately record the optimization time of pose graph optimization and bundle adjustment when the SLAM system detect loop in each sequence to compare the efficiency. Besides, we record the final pose estimation results of each frame, and calculate trajectory estimation RMSE as the comparison for accuracy .

The final experimental results are shown in Table \ref{table:Base Line}. Compared with the traditional optimization algorithm of OpenVSLAM, our method has almost no loss in terms of accuracy, but has greatly improved the efficiency. We observe unexpected slight improvement of accuracy of our method in some sequences. This may occur when the optimization result falls into local optimal as too many frames are involved in the optimization. More investigations are needed for the underlying cause.

\subsection{High-efficiency Optimization Experiment}
\label{high-effiency experiment}
In the comparison experiments of high-efficiency optimization algorithms, the dataset used is the 00-10 sequence of the KITTI datasets. We delete the 01 sequence because the visual SLAM system will lost during the tracking stage. In the experiment, our method is compared with the traditional SLAM algorithm ORBSLAM2\cite{mur2017orb}, and two best hierarchy optimization algorithms ENFT SfM\cite{zhang2016efficient} and Hybrid BA\cite{li2020hybrid}.

Since there are optimization algorithms suitable for SfM, in order to ensure the fairness of the experiment, we have adaptively modified the back-end optimization in the SLAM system. After the tracking of all frames in the SLAM system is completed, we will execute a global bundle adjustment again. And we record the final global bundle adjustment time as the evaluation for efficiency. At the same time, the Absolute Trajectory Error (ATE) of the final optimization result is used as the evaluation of accuracy. The experiment results are shown in Table \ref{table:high efficiency}. The results show that our method achieves the best performance in segment-based optimization algorithm both in terms of accuracy and efficiency.

\begin{table}[tp]
\setlength{\tabcolsep}{0.7mm}{
\begin{center}
\caption{\textbf{High Efficiency Optimization Experiment.} The results shown in this table are RMSE \textbf{/} Time and the units are respectively meter(m) and second(s). Note that in order to ensure fairness, we record the final global bundle adjustment time after the completion of tracking as the evaluation for efficiency. Absolute Trajectory Error (ATE) of the final optimization result is
used as the evaluation of accuracy. ACR stands for average change rate compared with the first column.}
\label{table:high efficiency}
\begin{tabular}{c|cccc|c}
\hline
KITTI & \tabincell{c}{ORB\\SLAM2\\{[9]}} & \tabincell{c}{ENFT\\SFM\\{[1]}} & \tabincell{c}{Hybyrid\\BA(ORB)\\{[3]}}  & \tabincell{c}{Hybyrid\\BA(DL)\\{[3]}} & \tabincell{c}{Our\\Method\\} \\
\hline \hline
00 & 1.43/11.7 & 4.80/8.25 & 4.90/\textbf{3.42} & 2.80/8.39 & \textbf{1.33}/4.16\\
02 & 5.72/19.6 & 28.3/14.7 & 18.2/4.43 & 15.9/13.2 & \textbf{5.72}/\textbf{3.94}\\
03 & 0.71/2.57 & 2.90/1.79 & 1.50/0.78 & 1.20/1.44 & \textbf{0.67}/\textbf{0.30}\\
04 & \textbf{0.20}/0.97 & 0.70/0.87 & 0.90/0.14 & 0.30/0.59 & 0.21/\textbf{0.21}\\
05 & \textbf{0.82}/8.77 & 3.50/6.18 & 2.90/4.87 & 3.20/6.23 & 0.85/\textbf{2.60}\\
06 & 0.83/4.60 & 14.4/2.53 & 8.20/0.65 & 9.20/3.75 & \textbf{0.82}/\textbf{1.15}\\
07 & 0.57/2.69 & 2.00/0.67 & 2.20/0.70 & 2.20/1.54 & \textbf{0.52}/\textbf{0.65}\\
08 & \textbf{3.65}/22.9 & 28.3/11.5 & 33.6/10.1 & 21.9/21.4 & 3.68/\textbf{3.26}\\
09 & \textbf{1.98}/4.65 & 5.90/2.04 & 6.50/1.96 & 6.20/5.57 & 2.04/\textbf{1.34}\\
10 & \textbf{1.04}/3.15 & 18.5/2.62 & 7.60/0.69 & 6.00/2.12 & 1.19/\textbf{0.67}\\
\hline

\hline
\tabincell{c}{ACR} & +0\%/+0\% & +595\%/-37\% & +403\%/-70\% & +317\%/-25\% & \textbf{+0.4\%}/\textbf{-77\%} \\
\hline 

\hline

\end{tabular}
\end{center}
\vspace{-0.5cm}
}
\end{table}
\begin{table}[tp]
\caption{\textbf{Ablation Experiment.} The results shown in this table are RMSE \textbf{/} Time and the units are respectively meter(m) and second(s). RMSE is used to measure accuracy and time is used to measure efficiency.}
\vspace{-0.2cm}
\label{table:Ablation}
\centering
\subtable[Segment Method]{
\begin{tabular}{c|cccc|c}
\hline
KITTI & \tabincell{c}{Reproj\\error} & Velocity & Covisibility  & \tabincell{c}{Fixed\\Length} & \tabincell{c}{Our\\Method} \\
\hline \hline
00 & 1.77/\textbf{4.02} & 1.89/3.98 & 1.39/7.17 & 1.48/4.77 & \textbf{1.33}/4.16\\
02 & 5.98/\textbf{3.72} & 6.24/3.77 & \textbf{5.71}/6.72 & 6.35/4.86 & 5.72/3.94\\
03 & 0.97/0.31 & 1.27/0.30 & 0.71/0.53 & 1.25/0.50 & \textbf{0.67}/\textbf{0.30}\\
04 & 0.23/0.21 & 0.29/0.23 & 0.25/0.35 & 0.27/0.28 & \textbf{0.21}/\textbf{0.21}\\
05 & 1.01/2.43 & 1.17/\textbf{1.99} & 0.88/2.96 & 1.23/2.96 & \textbf{0.85}/2.60\\
06 & 0.86/1.17 & 0.93/\textbf{1.13} & \textbf{0.79}/1.46 & 0.97/0.87 & 0.82/1.15\\
07 & 0.62/0.59 & 0.66/\textbf{0.51} & 0.61/0.98 & 0.77/0.80 & \textbf{0.52}/0.65\\
08 & 3.79/\textbf{3.03} & 3.98/3.43 & 3.72/3.59 & 3.98/3.50 & \textbf{3.68}/3.26\\
09 & 2.47/1.17 & 2.44/1.33 & 2.12/1.77 & 2.29/1.66 & \textbf{2.04}/\textbf{1.33}\\
10 & 1.24/\textbf{0.55} & 1.28/0.59 & 1.23/0.72 & 1.45/0.97 & \textbf{1.19}/0.67\\
\hline
\end{tabular}
\label{table:Segment}
}
\setlength{\tabcolsep}{1.39mm}{
\subtable[Buffer Method]{        
\begin{tabular}{c|cc}
\hline
KITTI & \tabincell{c}{Without\\Buffer} & Buffer \\
\hline \hline
00 & 1.84/3.17 & 1.33/4.16\\
02 & 6.98/2.98 & 5.72/3.94\\
03 & 0.75/0.24 & 0.67/0.30\\
04 & 0.36/0.15 & 0.21/0.21\\
05 & 0.94/2.27 & 0.85/2.60\\
06 & 1.27/1.04 & 0.82/1.15\\
07 & 0.77/0.61 & 0.52/0.65\\
08 & 3.98/3.29 & 3.68/3.26\\
09 & 2.59/1.37 & 2.04/1.34\\
10 & 1.73/0.64 & 1.19/0.67\\
\hline
\end{tabular}
\label{table:Buffer}
}\subtable[Interpolation Method]{
\begin{tabular}{c|cc|c}
\hline
& \tabincell{c}{Local\\BA} & \tabincell{c}{Linear\\Interpolation} & \tabincell{c}{Our\\Method} \\
\hline \hline
00 & 1.32/11.7 & 2.74/\textbf{4.07} & \textbf{}1.32/4.16\\
02 & \textbf{}5.71/12.5 & 6.65/3.87 & 5.72/3.94\\
03 & 0.69/2.24 & 1.14/\textbf{0.28} & \textbf{0.67}/0.30\\
04 & \textbf{0.19}/0.87 & 0.65/\textbf{0.15} & 0.21/0.21\\
05 & \textbf{0.84}/6.54 & 1.15/\textbf{2.43} & 0.85/2.60\\
06 & \textbf{0.81}/3.33 & 1.57/\textbf{1.02} & 0.82/1.15\\
07 & \textbf{0.53}/2.43 & 1.52/\textbf{0.59} & 0.52/0.65\\
08 & \textbf{3.63}/11.0 & 8.88/\textbf{2.87} & 3.68/3.26\\
09 & \textbf{1.99}/4.12 & 7.79/\textbf{1.03} & 2.04/1.34\\
10 & 1.21/3.01 & 5.87/\textbf{}0.45 & \textbf{1.19}/0.67\\
\hline
\end{tabular}
\label{table:Interpolation}
}
}
\vspace{-0.8cm}
\end{table}
\subsection{Ablation Experiment}
In order to verify the effect of each module in our proposed method, we conduct the ablation experiments. The experimental method is similar to that in Section \ref{high-effiency experiment}. A comparison experiment is executed on 00-10 sequences of KITTI. The efficiency and accuracy of the final global bundle adjustment are recorded for comparison. The results are shown in Table \ref{table:Ablation}.

Ablation experiments are divided into three parts, which are respectively used to verify the effect of our segmentation method, buffer mechanism and interpolation method.

In the first experiment, in order to verify the superiority of our segmentation method, we compare the proposed segmentation method with several traditional segmentation methods, including number of common viewpoints, reprojection error, speed and fixed-length segmentation. It is verified that our segmentation method has better performance both in terms of accuracy and efficiency in optimization. 

Then in order to verify the effect of the buffer, we remove the buffer area and change to direct connection between segments while other improvements remain. The experimental results show that after deleting the buffer, it shows a decline in the accuracy and efficiency of the optimization. 

Finally, we compare our interpolation method with the method using local BA and linear interpolation to verify the effectiveness of our interpolation method. And the results also show that our interpolation method perform the best.

\section{Conclusion}

In this paper, we propose a hierarchical segment-based optimization method in SLAM system. We propose a reliable trajectory segmentation method that can be used to increase efficiency in the back-end optimization. Then we propose a buffer mechanism for the first time to improve the robustness of the segmentation. Besides, we use global information to optimize the frames with poor estimation, and interpolation instead of optimization to update well-estimated frames. This hierarchical optimization method can greatly improve efficiency while ensuring that there is almost no loss in accuracy and outperforms the previous high-efficiency optimization method. And the comparative experiments also verify the effectiveness of our method. However, there are still some improvement room for hierarchical segmentation optimization including using more theoretical methods to replace the current segmentation method or proposing tightly-coupled method of segmentation and optimization to achieve better results.

\addtolength{\textheight}{-12cm}
\bibliographystyle{IEEEtran} 
\bibliography{refs} 

\begin{thebibliography}{10}
\providecommand{\url}[1]{#1}
\csname url@samestyle\endcsname
\providecommand{\newblock}{\relax}
\providecommand{\bibinfo}[2]{#2}
\providecommand{\BIBentrySTDinterwordspacing}{\spaceskip=0pt\relax}
\providecommand{\BIBentryALTinterwordstretchfactor}{4}
\providecommand{\BIBentryALTinterwordspacing}{\spaceskip=\fontdimen2\font plus
\BIBentryALTinterwordstretchfactor\fontdimen3\font minus
  \fontdimen4\font\relax}
\providecommand{\BIBforeignlanguage}[2]{{%
\expandafter\ifx\csname l@#1\endcsname\relax
\typeout{** WARNING: IEEEtran.bst: No hyphenation pattern has been}%
\typeout{** loaded for the language `#1'. Using the pattern for}%
\typeout{** the default language instead.}%
\else
\language=\csname l@#1\endcsname
\fi
#2}}
\providecommand{\BIBdecl}{\relax}
\BIBdecl

\bibitem{ni2007out}
K.~Ni, D.~Steedly, and F.~Dellaert, ``Out-of-core bundle adjustment for
  large-scale 3d reconstruction,'' in \emph{2007 IEEE 11th International
  Conference on Computer Vision}.\hskip 1em plus 0.5em minus 0.4em\relax IEEE,
  2007, pp. 1--8.

\bibitem{fang2019merge}
M.~Fang, T.~Pollok, and C.~Qu, ``Merge-{S}f{M}: {M}erging {P}artial
  {R}econstructions.'' in \emph{BMVC}, 2019, p.~29.

\bibitem{wu2011multicore}
C.~Wu, S.~Agarwal, B.~Curless, and S.~M. Seitz, ``Multicore bundle
  adjustment,'' in \emph{CVPR 2011}.\hskip 1em plus 0.5em minus 0.4em\relax
  IEEE, 2011, pp. 3057--3064.

\bibitem{eriksson2016consensus}
A.~Eriksson, J.~Bastian, T.-J. Chin, and M.~Isaksson, ``A consensus-based
  framework for distributed bundle adjustment,'' in \emph{Proceedings of the
  IEEE Conference on Computer Vision and Pattern Recognition}, 2016, pp.
  1754--1762.

\bibitem{li2020hybrid}
X.~Li and H.~Ling, ``{H}ybrid {C}amera{P}ose {E}stimation with {O}nline
  {P}artitioning for {SLAM},'' \emph{IEEE Robotics and Automation Letters},
  vol.~5, no.~2, pp. 1453--1460, 2020.

\bibitem{zhou2020stochastic}
L.~Zhou, Z.~Luo, M.~Zhen, T.~Shen, S.~Li, Z.~Huang, T.~Fang, and L.~Quan,
  ``Stochastic bundle adjustment for efficient and scalable 3d
  reconstruction,'' in \emph{European Conference on Computer Vision}.\hskip 1em
  plus 0.5em minus 0.4em\relax Springer, 2020, pp. 364--379.

\bibitem{zhang2016efficient}
G.~Zhang, H.~Liu, Z.~Dong, J.~Jia, T.-T. Wong, and H.~Bao, ``Efficient
  non-consecutive feature tracking for robust structure-from-motion,''
  \emph{IEEE Transactions on Image Processing}, vol.~25, no.~12, pp.
  5957--5970, 2016.

\bibitem{hierarchy2014}
B.~Suger, G.~D. Tipaldi, L.~Spinello, and W.~Burgard, ``An approach to solving
  large-scale slam problems with a small memory footprint,'' in \emph{2014 IEEE
  International Conference on Robotics and Automation (ICRA)}.\hskip 1em plus
  0.5em minus 0.4em\relax IEEE, 2014, pp. 3632--3637.

\bibitem{KITTI}
A.~Geiger, P.~Lenz, and R.~Urtasun, ``{A}re we ready for {A}utonomous
  {D}riving? {T}he {KITTI} {V}ision {B}enchmark {S}uite,'' in \emph{Conference
  on Computer Vision and Pattern Recognition (CVPR)}, 2012.

\bibitem{tum-rgbd}
J.~Sturm, N.~Engelhard, F.~Endres, W.~Burgard, and D.~Cremers, ``{A}
  {B}enchmark for the {E}valuation of {RGB-D SLAM} {S}ystems,'' in \emph{Proc.
  of the International Conference on Intelligent Robot Systems (IROS)}, Oct.
  2012.

\bibitem{euroc}
``The {EuRoC} micro aerial vehicle datasets,'' \emph{The International Journal
  of Robotics Research}, vol.~35, no.~10, pp. 1157--1163, 2016.

\bibitem{engels2006bundle}
C.~Engels, H.~Stew{\'e}nius, and D.~Nist{\'e}r, ``Bundle adjustment rules,''
  \emph{Photogrammetric computer vision}, vol.~2, no.~32, 2006.

\bibitem{kummerle2011g}
R.~K{\"u}mmerle, G.~Grisetti, H.~Strasdat, K.~Konolige, and W.~Burgard, ``g 2
  o: A general framework for graph optimization,'' in \emph{2011 IEEE
  International Conference on Robotics and Automation}.\hskip 1em plus 0.5em
  minus 0.4em\relax IEEE, 2011, pp. 3607--3613.

\bibitem{kaess2008isam}
M.~Kaess, A.~Ranganathan, and F.~Dellaert, ``i{SAM}: {I}ncremental smoothing
  and mapping,'' \emph{IEEE Transactions on Robotics}, vol.~24, no.~6, pp.
  1365--1378, 2008.

\bibitem{liu2018ice}
H.~Liu, M.~Chen, G.~Zhang, H.~Bao, and Y.~Bao, ``Ice-ba: {I}ncremental,
  consistent and efficient bundle adjustment for visual-inertial slam,'' in
  \emph{Proceedings of the IEEE Conference on Computer Vision and Pattern
  Recognition}, 2018, pp. 1974--1982.

\bibitem{ila2017fast}
V.~Ila, L.~Polok, M.~Solony, and K.~Istenic, ``Fast incremental bundle
  adjustment with covariance recovery,'' in \emph{2017 International Conference
  on 3D Vision (3DV)}.\hskip 1em plus 0.5em minus 0.4em\relax IEEE, 2017, pp.
  175--184.

\bibitem{indelman2012incremental}
V.~Indelman, R.~Roberts, C.~Beall, and F.~Dellaert, ``Incremental light bundle
  adjustment.''\hskip 1em plus 0.5em minus 0.4em\relax Georgia Institute of
  Technology, 2012.

\bibitem{vo2016spatiotemporal}
M.~Vo, S.~G. Narasimhan, and Y.~Sheikh, ``Spatiotemporal bundle adjustment for
  dynamic 3d reconstruction,'' in \emph{Proceedings of the IEEE Conference on
  Computer Vision and Pattern Recognition}, 2016, pp. 1710--1718.

\bibitem{schonberger2016structure}
J.~L. Schonberger and J.-M. Frahm, ``Structure-from-motion revisited,'' in
  \emph{Proceedings of the IEEE conference on computer vision and pattern
  recognition}, 2016, pp. 4104--4113.

\bibitem{shah2015multistage}
R.~Shah, A.~Deshpande, and P.~Narayanan, ``{M}ultistage {S}f{M}: A
  coarse-to-fine approach for 3d reconstruction,'' \emph{arXiv preprint
  arXiv:1512.06235}, 2015.

\bibitem{suger2014approach}
B.~Suger, G.~D. Tipaldi, L.~Spinello, and W.~Burgard, ``An approach to solving
  large-scale slam problems with a small memory footprint,'' in \emph{2014 IEEE
  International Conference on Robotics and Automation (ICRA)}.\hskip 1em plus
  0.5em minus 0.4em\relax IEEE, 2014, pp. 3632--3637.

\bibitem{jang2020pose}
Y.~Jang, H.~Shin, and H.~J. Kim, ``{P}ose {C}orrection {A}lgorithm for
  {R}elative {F}rames between {K}eyframes in {SLAM},'' in \emph{Proceedings of
  the Asian Conference on Computer Vision}, 2020.

\bibitem{sumikura2019openvslam}
S.~Sumikura, M.~Shibuya, and K.~Sakurada, ``{O}pen{VSLAM}: {A} versatile visual
  {SLAM} framework,'' in \emph{Proceedings of the 27th ACM International
  Conference on Multimedia}, 2019, pp. 2292--2295.

\bibitem{mur2017orb}
R.~Mur-Artal and J.~D. Tard{\'o}s, ``Orb-slam2: An open-source slam system for
  monocular, stereo, and rgb-d cameras,'' \emph{IEEE Transactions on Robotics},
  vol.~33, no.~5, pp. 1255--1262, 2017.

\end{thebibliography}
\end{document}